# Optimizing Multi-Class Text Classification: A Diverse Stacking Ensemble Framework Utilizing Transformers

Anusuya Krishnan*

**Abstract:** Customer reviews play a crucial role in assessing customer satisfaction, gathering feedback, and driving improvements for businesses. Analyzing these reviews provides valuable insights into customer sentiments, including compliments, comments, and suggestions. Text classification techniques enable businesses to categorize customer reviews into distinct categories, facilitating a better understanding of customer feedback. However, challenges such as overfitting and bias limit the effectiveness of a single classifier in ensuring optimal prediction. This study proposes a novel approach to address these challenges by introducing a stacking ensemble-based multi-text classification method that leverages transformer models. By combining multiple single transformers, including BERT, ELECTRA, and DistilBERT, as base-level classifiers, and a meta-level classifier based on RoBERTa, an optimal predictive model is generated. The proposed stacking ensemble-based multi-text classification method aims to enhance the accuracy and robustness of customer review analysis. Experimental evaluations conducted on a real-world customer review dataset demonstrate the effectiveness and superiority of the proposed approach over traditional single classifier models. The stacking ensemble-based multi-text classification method using transformers proves to be a promising solution for businesses seeking to extract valuable insights from customer reviews and make data-driven decisions to enhance customer satisfaction and drive continuous improvement.

## 1 Introduction

In the digital era, the significance of customer reviews cannot be overstated in shaping the prosperity and image of businesses spanning diverse industries. These reviews serve as invaluable feedback and insights, acting as a vital informational resource for organizations to assess customer satisfaction, pinpoint areas for enhancement, and drive data-informed decision-making.

The analysis of customer reviews has gained immense importance as businesses endeavor to refine their offerings, improve service quality, and elevate the overall customer experience[1].

The ability to automatically classify text data into predefined categories is crucial in modern information processing systems. Text classification, a fundamental task in natural language processing (NLP), plays a crucial role in organizing, categorizing, and extracting insights from textual data. One of the most prevalent forms of text classification is multi-class text classification, where a given piece of text is assigned to one of several predefined categories or classes.

Multi-class text classification finds application in various real-world scenarios, spanning across different domains and industries. From sentiment analysis, where texts are categorized into positive, negative, or

● Anusuya Krishnan is with the Data science research group in College of information technology, United Arab Emirates University, UAE. E-mail: anusuyababy18@gmail.com

∗ To whom correspondence should be addressed.

Manuscript received: year-month-day; accepted: year-month-day

neutral sentiments, to topic categorization in news articles and customer support ticket classification, multi-class text classification enables efficient and automated handling of textual data at scale. The goal of multi-class text classification is to build predictive models that can accurately and efficiently classify text documents into multiple classes based on their content and context. This poses unique challenges compared to binary or single-class text classification, as the number of potential classes increases, leading to class imbalances and increased model complexity[2].

To tackle these challenges, researchers and practitioners in the NLP community have explored various machine learning techniques and deep learning approaches. Traditional methods, such as Naive Bayes, Support Vector Machines (SVM), and Random Forests, have been employed with feature engineering techniques to extract relevant information from text data.

However, in recent years, transformer-based models have demonstrated remarkable success in NLP tasks, thanks to their attention mechanisms and ability to capture contextual information effectively. Models such as BERT (Bidirectional Encoder Representations from Transformers), ELECTRA (Efficiently Learning an Encoder that Classifies Token Replacements Accurately), DistilBERT (Distilled Bidirectional Encoder Representations from Transformers) and RoBERTa (A Robustly Optimized Bidirectional Encoder Representations from Transformers Pretraining Approach) have achieved state-of-the-art results in various benchmarks[22-25]. Despite their capabilities, utilizing transformer models in multi-class text classification presents certain challenges.

One of the key challenges is the limited perspective each individual transformer model provides. Each model may be biased towards certain linguistic patterns or contextual clues, leading to suboptimal predictions on specific instances within the dataset. Moreover, individual models may struggle to handle class imbalances and noisy data, affecting their overall performance.

To address these limitations, we propose the "Diverse Stacking Ensemble Framework" for optimizing multi-class text classification. Our framework leverages the strength of ensemble learning and diversity in transformer architectures to create a powerful and robust classification system. Instead of relying on a single transformer model, we combine the predictions from multiple transformer models, each with distinct pretrained weights or architectures. The ensemble approach allows us to capture a more comprehensive understanding of the text data by leveraging the complementary strengths of different transformer models. By doing so, we aim to reduce the impact of biases and limitations inherent in individual models, leading to more accurate and reliable predictions.

In this study, we present a detailed description of our proposed Diverse Stacking Ensemble Framework for multi-class text classification. We outline the steps involved in preparing the dataset, preprocessing the text data, and selecting diverse transformer models. We also describe the ensemble technique used to combine the predictions from these models effectively. To validate the effectiveness of our framework, we conduct extensive experiments on benchmark datasets and compare the performance of our ensemble model against single transformer models and traditional machine learning classifiers. We demonstrate the significant improvements achieved by the ensemble approach, especially in scenarios with imbalanced classes and noisy data.

The contributions of this paper are twofold: firstly, we propose a novel Diverse Stacking Ensemble Framework for multi-class text classification, showcasing the advantages of leveraging diverse transformer models in ensemble learning; secondly, we present empirical evidence supporting the superiority of our framework over conventional single-model approaches. In the subsequent sections, we provide a comprehensive explanation of our framework's design, implementation, and empirical results, demonstrating its potential to advance the state-of-the-art in multi-class text classification using transformer-based models.

## 2 Related work

In recent years, there has been a notable increase in research dedicated to the field of multi text classification for customer reviews. Scholars have extensively explored different techniques, models, and applications, shedding light on the advantages and hurdles associated with employing text classification for analyzing customer feedback. This growing attention stems from the necessity to efficiently analyze vast amounts of text data and unveil underlying themes and patterns within it. Researchers have ventured into diverse methodologies, merging the capabilities of machine learning with multi text classification techniques to enhance their analyses [1]. Over the years, numerous studies have focused on developing effective techniques and models to tackle the complexities of this task, aiming to improve accuracy, efficiency, and scalability. We explore some of the key advancements and approaches proposed by researchers in the field of multi-class text classification[2-5].

Early research in multi-class text classification often employed traditional machine learning algorithms such as Naive Bayes, Support Vector Machines (SVM), and Random Forests. These algorithms were combined with various feature extraction techniques, including bag-of-words, TF-IDF, and n-grams. While these approaches achieved moderate success, they often faced challenges in capturing complex semantic relationships and handling large-scale textual data effectively[6-7]. In the past, researchers employed traditional machine learning methods, such as Support Vector Regression (SVR) to measure the similarity between texts[8,33]. However, in recent times, numerous deep learning techniques have emerged, including pre-trained models, offering more advanced approaches for this task[10].

The rise of neural network-based models revolutionized the field of NLP, enabling the development of more sophisticated and powerful classifiers. Feedforward neural networks and recurrent neural networks (RNNs) were among the early neural models used for text classification. However, they struggled with vanishing gradients and the inability to handle long-range dependencies in text. CNNs emerged as a breakthrough in multi-class text classification due to their ability to learn hierarchical features from text data[11-15]. The author introduced a CNN architecture for text classification, where one-dimensional convolutions were applied over word embeddings to capture local features effectively. CNNs demonstrated promising results and became widely adopted for various NLP tasks[7-9].

Another study proposed a sentiment analysis framework utilizing the Doc2Vec word embedding model for feature vector representation. They employed seven different classifiers, namely K nearest neighbor (KNN), AdaBoost, support vector machine (SVM), decision tree, random forest, logistic regression, and naïve Bayes, for classifying the embedded feature vectors. Evaluating their framework on the US airline services dataset, they achieved the highest accuracy of 84.5% using the ensemble-based AdaBoost classifier. However, the limited size of the dataset could lead to underfitting issues, affecting generalization to new, unseen data instances[29].

In contrast, researchers explored the use of Word2Vec and GLoVe word embedding techniques, along with the LSTM model, for sentiment identification in US airlines' tweets dataset[20-23]. They categorized tweets into positive, neutral, and negative sentiments[16, 22]. Despite achieving an accuracy of around 75%, their validation curve plateaued, indicating limitations in generalizing to new data instances. The authors explored the use of fusion models, combining CNN, LSTM, and GRU architectures on the IMDB review dataset, to enhance the performance of baseline LSTM and GRU-based frameworks[17]. Their findings revealed a significant accuracy improvement, from 85% to 89%, by employing fusion models. However, this enhanced accuracy came at the expense of increased computational overhead, making it computationally expensive.

Word embeddings, such as Word2Vec and GloVe played a crucial role in advancing multi-class text classification. These methods transformed words into dense vector representations, capturing semantic similarities between words[16, 18-21]. Word embeddings provided valuable contextual information, enhancing the performance of traditional machine learning algorithms and neural networks. RNNs and LSTM networks were also popular choices for multi-class text classification tasks, especially for tasks involving sequential data like sentiment analysis and named entity recognition. LSTMs, in particular, addressed the vanishing gradient problem and enabled the capture of long-range dependencies in text. The advent of pretrained language models marked a significant turning point in multi-class text classification. Models like BERT and RoBERTa demonstrated state-of-the-art performance across various NLP tasks, including text classification[22-23]. These transformer-based models leverage attention mechanisms and large-scale pretraining on vast text corpora to capture rich contextual information and semantic relationships in text. Researchers explored the fine-tuning of pretrained language models for multi-class text classification tasks[24]. Devlin et al. (2018) showed that by adding a simple linear layer on top of BERT and fine-tuning it on the specific classification task, impressive results could be achieved. Fine-tuning proved to be an efficient transfer learning technique, allowing the models to adapt to new tasks with limited labeled data[25-27].

## 3  Methodology

In this section, a thorough explanation of the multi-class text classification framework is presented, which is built upon a novel heterogeneous stacking ensemble approach. The success of pre-trained models in Natural Language Processing (NLP) owes much to the incorporation of attention mechanisms. Transformer-based models like BERT, RoBERTa, ELECTRA, and DistilBERT have played a significant role in producing effective hidden representations. To further improve performance, we adopted a stacking ensemble strategy to combine and leverage the results obtained from these models. The overall architecture of the proposed framework can be observed in Figure 2. Within this figure, the framework is composed of several sub-modules, each dedicated to performing specific tasks. Elaborate descriptions of these sub-modules are provided below.

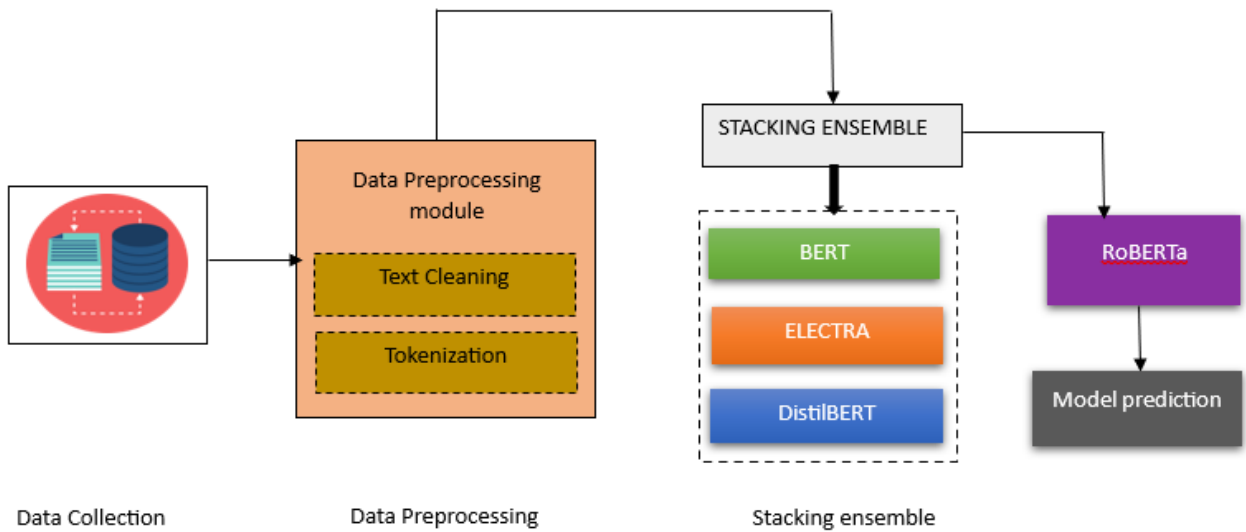

Fig. 1    Architecture of the proposed diagram

## 3.1 Data Collection

This study employed two distinct datasets: the OpSpam dataset and the Customer Satisfaction dataset from the UAE Ministry of Economy (MOE) government website.

The first dataset consists of 29,200 reviews acquired from the UAE Ministry of Economy. These reviews are genuine and represent the levels of customer satisfaction with the MOE app. They offer valuable insights into real user experiences and opinions concerning the government's services.

The second dataset is the publicly available OpSpam dataset, accessible through Kaggle. It comprises 1600 customer reviews collected from prominent platforms like TripAdvisor and Amazon Mechanical Turk. This dataset serves as a valuable resource for sentiment analysis and opinion mining tasks, as it offers a diverse range of reviews suitable for analysis and research purposes. Researchers and practitioners can utilize this dataset to delve into customer opinions, sentiment trends, and other relevant topics across various domains.

## 3.2 Data preprocessing

The data preprocessing module plays a crucial role in refining and cleaning the raw data, ensuring that it contains only the necessary features for the topic modelling task. This module employs various techniques to enhance the quality of the data. It effectively removes irrelevant phrases and symbols, eliminating any noise or distractions from the text. Additionally, stopword elimination is applied to remove commonly used words in the language that do not contribute significantly to text mining tasks. These words, such as prepositions, numbers, and other irrelevant terms, lack relevant information for the study.

To facilitate deep analysis, tokenization is employed, dividing the text input into meaningful units such as phrases, words, or tokens. The outcome of tokenization is a sequence of these tokens, which serve as the fundamental units for further analysis and processing. Furthermore, lemmatization with part-of-speech (POS) tagging is applied to optimize the data. This process reduces the feature space by mapping words to their base or dictionary forms, reducing redundancy and improving the efficiency of subsequent analyses.

## 3.3 Stacking ensemble transformer models

This module incorporates several natural language processing models, including BERT, ELECTRA, and DistilBERT, as base-level classifiers. Additionally, it employs a Roberta-based meta-level classifier. The details of each transformer are represented as below.

### 3.3.1 BERT

Bidirectional Encoder Representations from Transformers (BERT) is designed to pretrain deep bidirectional representations from unlabeled text by considering both left and right context in all layers. We used the "BERT-base-multilingual-cased" checkpoint, which comprises 12 layers, 768 hidden units, 12 heads, and a total of 110 million parameters. The model was trained on cased text in the top 104 languages with the largest Wikipedias. The output tensor contains batch_size, sequence_length, and hidden_state, with the first token utilized for regression[21, 23]. Additionally, BERT uses character-level BPE encoding.

### 3.3.2 ELECTRA

ELECTRA, short for "Efficiently Learning an Encoder that Classifies Token Replacements Accurately," is another powerful language model that aims to improve pretraining efficiency. Similar to BERT, ELECTRA is designed to pretrain deep contextualized representations from unlabeled text. However, it employs a novel approach known as "replaced token detection" instead of traditional masked language modeling. We used the "ELECTRA-base" checkpoint for our implementation, which consists of 12 layers, 768 hidden units, 12 attention heads, and a total of 110 million parameters. Like BERT, ELECTRA was also trained on a diverse dataset, but it employs its unique replaced token detection approach during pretraining.

The output tensor from ELECTRA includes batch_size, sequence_length, and hidden_state, just like BERT. The first token of the hidden_state can be utilized for specific downstream tasks, such as regression or classification. Additionally, ELECTRA relies on character-level Byte Pair Encoding (BPE) for tokenization, ensuring effective representation of characters in the text.

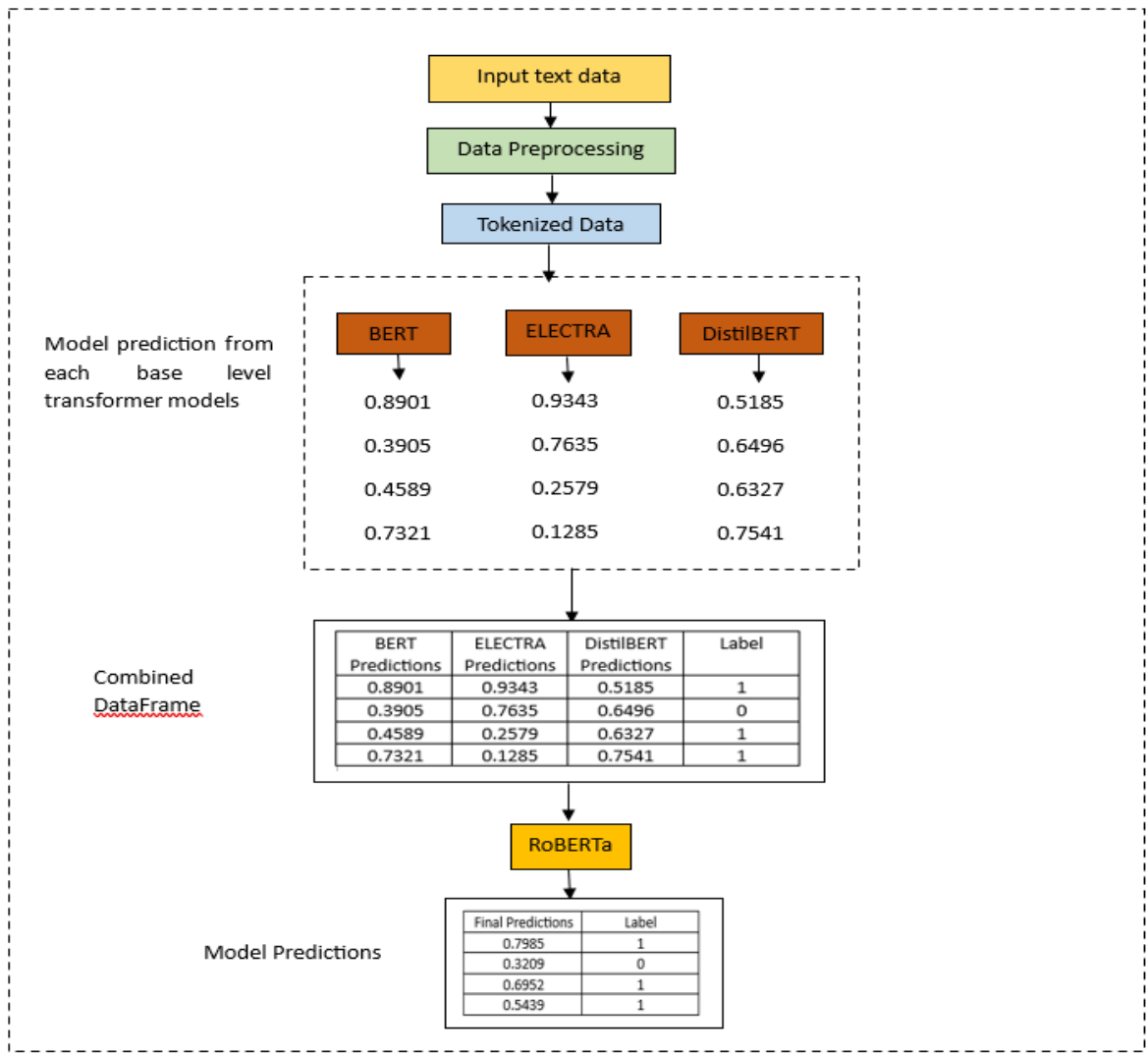

Fig. 2    Schematic representation of visualizing the training Process in an ensemble-based framework

### 3.3.3 DistilBERT

DistilBERT is a distilled version of BERT, which involves pre-training a smaller general-purpose language representation model that can be fine-tuned to achieve good performances across various tasks, similar to its larger counterparts. The main idea behind DistilBERT is to employ knowledge distillation, where the knowledge from the larger BERT model is transferred to a smaller architecture. DistilBERT is constructed with 6 layers, 768 hidden units, 12 attention heads, and approximately 66 million parameters, making it much more lightweight compared to BERT, which typically contains 12 layers and around 110 million parameters. The output tensor from DistilBERT includes batch_size, sequence_length, and hidden_state, following a similar format to BERT and ELECTRA. It retains the ability to represent contextualized information effectively, making it well-suited for various natural language processing tasks.

### 3.3.4 RoBERTa

RoBERTa is an enhanced pretraining method built upon BERT, with the primary aim of surpassing various post-BERT approaches in performance. The modifications made in RoBERTa are straightforward and involve: extending the model training duration with larger batches and an expanded dataset, eliminating the next sentence prediction objective, training on longer sequences, and incorporating a dynamically changing masking pattern during training. The RoBERTa model utilizes the "roberta-base" checkpoint, which consists of a 12-layer architecture, 768 hidden units, 12 attention heads, and 125 million parameters. It adopts a similar architecture to BERT-base, resulting in outputs comparable to BERT. To indicate the start of a sentence, RoBERTa uses a special token called "sos," and it employs byte-level BPE encoding for tokenization.

### 3.4 Model Training

This ensemble-based framework is represented schematically in Figure 2, illustrating the flow of the training process. The input to the base-level classifiers is the preprocessed tokenized text documents encoded as integers, generated by the "data Preprocessing" module. After training the base-level classifiers, they predict the probability values for new data instances, indicating the likelihood of belonging to a class label, such as "Class1" in binary classification. If the probability value is greater than or equal to 0.5, the base classifier assigns the data instance to "Class1"; otherwise, it is assigned to "Class0." For datasets with multiple class labels, each base classifier produces an array of dimensions corresponding to the number of class labels, and the predicted class is determined by the array index with the highest probability value. The predicted probabilities from the base classifiers are then combined into a single dataframe, which serves as the feature set for training the Roberta-based meta-level classifier. This classifier uses the information from the base-level classifiers to make a final prediction.

The proposed stacking ensemble-based sentiment analysis framework utilizes two loss functions for performing binary and multi-class classification, respectively. For binary classification, the framework employs the "binary cross-entropy" loss function in Eq. (1):

$$Loss_{Binary} = -1/N \sum [(P_I \log \grave{P}_i + (1-P_I) \log(1-\grave{P}_I))] \quad (1)$$

where $\grave{P}_i$ represents the probability value assigned by the classifier indicating whether the given data instance belongs to the 'i'$^{th}$ class label (positive or negative sentiment), $P_i$ is the actual class label of the data instance, and N is the total number of data instances. Similarly, for multi-class classification, the framework uses the "sparse categorical cross-entropy" loss function in Eq. (2):

$$Loss_{Multiclass} = -1/N \sum [(P_I \log \grave{P}_i)] \quad (2)$$

where $\grave{P}_i$ represents the probability value assigned by the classifier indicating the likelihood of the given data instance belonging to the 'i'$^{th}$ class label (multiple sentiment classes), $P_i$ is the actual class label of the data instance, and N is the total number of data instances.

### 3.5 Evaluation metrics

The performance evaluation of the proposed stacking ensemble-based sentiment analysis framework incorporates the following key performance metrics (precision, recall, F1-Score, and accuracy) for text classification in Eq. (3-6):

$$Precision = TP / (TP+FP) \quad (3)$$

$$Recall = TP / (TP+FN) \quad (4)$$

$$F1\text{-}Score = 2 \times Precision \times Recall/(Precision+Recall) \quad (5)$$

$$Accuracy = TP+TN/(TP+TN+FP+FN) \quad (6)$$

where TP represents true positive, FP represents false positive, TN represents true negative, and FN represents false negative. These metrics provide a comprehensive assessment of the framework's effectiveness in handling both binary and multi-class sentiment analysis tasks. Additionally, the mean squared error (MSE) is calculated using the following formula in Eq. (7):

$$MSE = 1/N \times \sum(N)(P_i - \grave{P}_i)^2 \quad (7)$$

where $P_i$ represents the predicted value, and $\grave{P}_i$ represents the ground-truth value. The MSE metric is commonly used to evaluate the accuracy of predictions by measuring the average squared difference between predicted and actual values.

## 4 Experimental Results and Discussion

In this section, we detail the experimental setup and share the findings of our study, specifically concentrating on the comparison of individual transformer performances and our proposed ensemble stacking approaches. We delve into the specifics, emphasizing the significant discoveries and insights derived from these comparisons. By scrutinizing the outcomes of these experiments, we gain a profound understanding of the performance and efficacy of our proposed approach.

### 4.1 Experiment 1: Comprehensive Analysis of Customer happiness dataset

In this section, we undertook the process of transforming the raw dataset into a structured format. The preparation of the data involved a meticulous cleaning procedure that encompassed the elimination of redundant characters, numbers, stopwords, and symbols. To enhance the significance of sentences, we executed lemmatization while employing POS tagging. Subsequently, the refined data was channeled into the TF-IDF vectorizer to facilitate numerical representation. This data was then subjected to five distinct machine learning techniques, and a comparative analysis was conducted on the six models: LSVM, RF, LR, PAC, LGBM, and GB.

**Table 1 Performance Comparison of different machine learning techniques in dataset 1**

| Machine learning techniques | Performance metrics | | | |
|---|---|---|---|---|
| | Accuracy | Precision | Recall | F1-Score |
| LSVM | 0.82 | 0.82 | 0.81 | 0.81 |
| LR | 0.81 | 0.81 | 0.81 | 0.81 |
| RF | 0.80 | 0.79 | 0.81 | 0.81 |
| LGBM | 0.83 | 0.82 | 0.83 | 0.83 |
| GB | 0.82 | 0.80 | 0.83 | 0.82 |
| PAC | 0.81 | 0.81 | 0.80 | 0.81 |

Table 1 provides the performance metrics across these methodologies, the linear support vector Machine (LSVM) emerged as a standout performer, attaining noteworthy results, including an accuracy of 0.82, precision of 0.82, recall of 0.81, and an F1-Score of 0.81. Conversely, the logistic regression (LR) methodology showcased consistent convergence across accuracy, precision, recall, and F1-Score, all stabilizing at 0.81. Amid the domain of ensemble models, the random forest (RF) architecture demonstrated an accuracy of 0.80, precision of 0.79, recall of 0.81, and an F1-Score of 0.81. In stark contrast, the light gradient boosting machine (LGBM) distinguished itself by achieving the most remarkable performance, recording impressive accuracy, precision, recall, and F1-Score values, all set at 0.83. Meanwhile, the gradient boosting (GB) strategy secured an accuracy of 0.82, precision of 0.80, recall of 0.83, and an F1-Score of 0.82. Ultimately, the passive aggressive classifier (PAC) generated results encompassing an accuracy of 0.81, precision of 0.81, recall of 0.80, and an F1-Score of 0.81. These findings collectively underscore the multifaceted nature of performance exhibited by the array of examined machine learning techniques.

**Table 2 Performance Comparison of different pre trained transformer techniques in dataset 1**

| Pre-trained Transformer techniques | Performance metrics | | | |
|---|---|---|---|---|
| | Accuracy | Precision | Recall | Loss |
| BERT | 0.86 | 0.89 | 0.83 | 0.39 |
| ELECTRA | 0.86 | 0.87 | 0.84 | 0.76 |
| DistilBERT | 0.85 | 0.80 | 0.89 | 0.67 |
| RoBERTa | 0.87 | 0.88 | 0.85 | 0.29 |
| Our method | 0.90 | 0.91 | 0.88 | 0.21 |

Displayed within Table 2, subsequent to the refinement of raw data, the data was directly input into transformer models. This tabulated presentation offers a comprehensive exploration of performance metrics associated with a range of pre-trained Transformer techniques. BERT, a prominent member of this category, attained noteworthy results, showcasing an accuracy of 0.86, precision of 0.89, recall of 0.83, and a loss value of 0.39. Similarly, ELECTRA yielded commendable outcomes, securing an accuracy of 0.86, precision of 0.87, recall of 0.84, and a loss of 0.76. In the context of DistilBERT, an accuracy rating of 0.85 was accompanied by precision, recall, and loss values of 0.80, 0.89, and 0.67 respectively. RoBERTa exhibited a competitive performance, registering an accuracy of 0.87, precision of 0.88, recall of 0.85, and a loss of 0.29. Notably, our proposed stacking ensemble (SE) transformer model emerged as a standout performer, showcasing superior performance metrics: an accuracy of 0.90, precision of 0.91, recall of 0.88, and a loss value of 0.21. These meticulously evaluated findings collectively underscore the efficacy and potential of these pre-trained transformer techniques in addressing intricate tasks. Figure 3 showcases the performances of the transformer models.

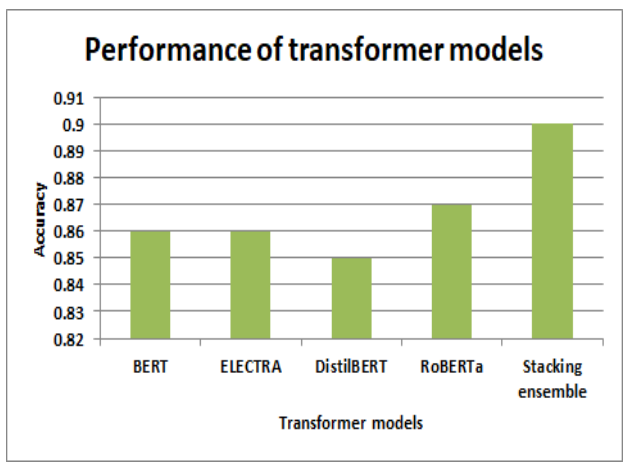

Fig. 3 Performance of transformer models on dataset 1

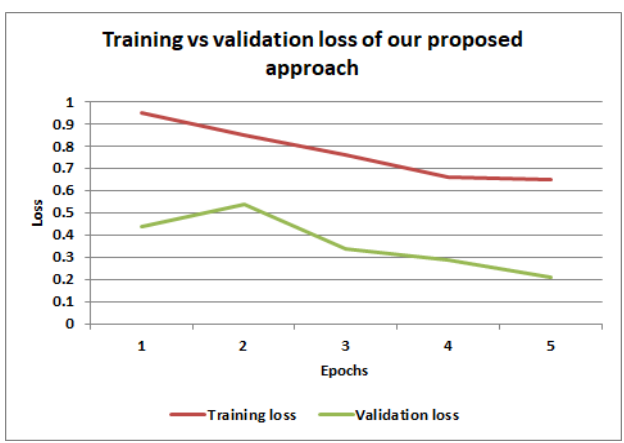

Fig. 4 Training and validation loss of stacking ensemble-based transformer model on dataset 1

Figure 4 illustrates the loss function value plots of the meta-level classifier, RoBERTa, employed within the proposed stacking ensemble framework. The analysis of this figure underscores that the validation losses of the meta-level classifier exhibit a plateau around 21% after the 5th iteration.

## 4.2 Experiment 2: Comprehensive Analysis of Hotel review dataset

In this segment, we replicated the methodology outlined in experiment 1. The process of data preparation involved a thorough cleaning routine, encompassing the removal of redundant characters, numbers, stopwords, and symbols. To elevate the importance of sentences, we employed lemmatization in combination with POS tagging. Following this step, the processed data was input into the TF-IDF vectorizer for numerical representation facilitation. Subsequently, this data underwent evaluation using five unique machine learning methods, and a comparative examination was performed on the six models (LSVM, RF, LR, PAC, LGBM, and GB).

Table 3 Performance Comparison of different machine learning techniques in dataset 1

| Machine learning techniques | Performance metrics | | | |
|---|---|---|---|---|
| | Accuracy | Precision | Recall | F1-Score |
| LSVM | 0.90 | 0.91 | 0.90 | 0.90 |
| LR | 0.89 | 0.86 | 0.91 | 0.89 |
| RF | 0.89 | 0.88 | 0.89 | 0.89 |
| LGBM | 0.88 | 0.86 | 0.90 | 0.88 |
| GB | 0.90 | 0.91 | 0.90 | 0.90 |
| PAC | 0.92 | 0.91 | 0.93 | 0.92 |

Table 3 employed encompassed accuracy, precision, recall, and F1-score, collectively providing a comprehensive perspective on the techniques' effectiveness. Among the methodologies explored, the linear support vector machine (LSVM) displayed a noteworthy accuracy of 0.90. The precision, recall, and F1-score associated with LSVM were observed to be 0.91, 0.90, and 0.90, respectively. Logistic regression (LR) demonstrated a commendable accuracy level of 0.89, with corresponding precision, recall, and F1-score values of 0.86, 0.91, and 0.89, respectively. Meanwhile, random forest (RF) yielded an accuracy of 0.89 and achieved precision, recall, and F1-score metrics of 0.88, 0.89, and 0.89, respectively, establishing its competitive performance. On the other hand, light gradient boosting machine (LGBM) exhibited an accuracy of 0.88, coupled with precision, recall, and F1-score measures of 0.86, 0.90, and 0.88, respectively, underscoring its proficiency within the range of techniques explored. The gradient boosting (GB) technique achieved an accuracy matching that of LSVM, reaching 0.90. Its precision, recall, and F1-score were recorded as 0.91, 0.90, and 0.90, respectively, solidifying its effectiveness. Ultimately, the passive aggressive classifier (PAC) outshone the other techniques, boasting an accuracy of 0.92. PAC's precision, recall, and F1-score were particularly impressive at 0.91, 0.93, and 0.92, respectively, rendering it the top-performing approach within this study's context.

**Table 4 Performance Comparison of different pre-trained transformer techniques in dataset 2**

| Pre-trained Transformer techniques | Performance metrics | | | |
|---|---|---|---|---|
| | Accuracy | Precision | Recall | Loss |
| BERT | 0.89 | 0.89 | 0.90 | 0.19 |
| ELECTRA | 0.88 | 0.88 | 0.89 | 0.26 |
| DistilBERT | 0.91 | 0.86 | 0.94 | 0.17 |
| RoBERTa | 0.92 | 0.90 | 0.93 | 0.16 |
| Our method | 0.94 | 0.94 | 0.93 | 0.09 |

Table 4 shows a comprehensive evaluation of diverse pre-trained Transformer techniques was conducted, with a focus on key performance metrics encompassing accuracy, precision, recall, and loss. Among these techniques, BERT displayed a notable accuracy of 0.89. Its precision, recall, and loss values were observed at 0.89, 0.90, and 0.19, respectively. Similarly, ELECTRA exhibited an accuracy of 0.88, accompanied by precision, recall, and loss metrics of 0.88, 0.89, and 0.26, respectively. DistilBERT demonstrated robust performance with an accuracy of 0.91, while achieving precision, recall, and loss scores of 0.86, 0.94, and 0.17, respectively. RoBERTa showcased noteworthy outcomes, boasting an accuracy of 0.92, and precision, recall, and loss figures of 0.90, 0.93, and 0.16, respectively. However, emerging as a frontrunner, our proposed method achieved exceptional results, with an accuracy of 0.94. Notably, this method displayed remarkable precision, recall, and loss measures, recording values of 0.94, 0.93, and 0.09, respectively. This rigorous evaluation provides valuable insights into the comparative performance of various pre-trained Transformer techniques. Figure 5 explains the performances of the transformer models.

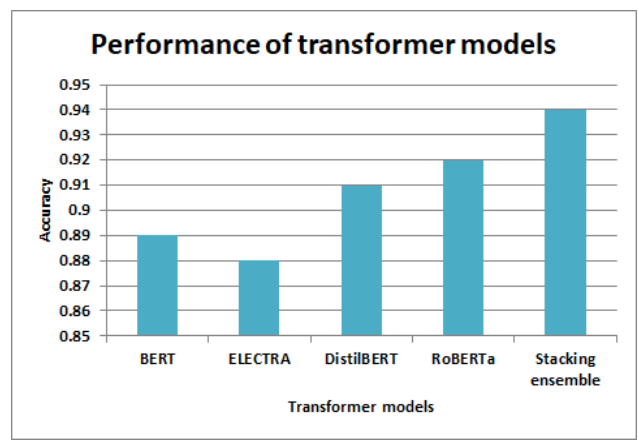

**Fig. 5 Performance of transformer models on dataset 2**

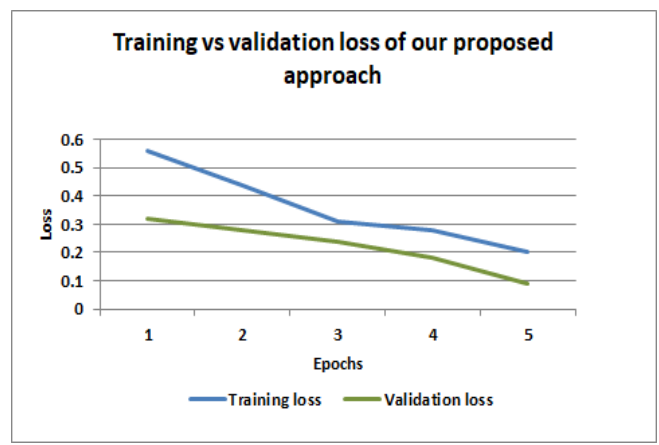

**Fig. 6 Training and validation loss of stacking ensemble-based transformer model on dataset 2**

In Figure 6, the plots depicting the values of the loss function for the meta-level classifier, RoBERTa, utilized in the proposed stacking ensemble framework are presented. The examination of this visual representation highlights a notable observation – the validation losses of the meta-level classifier reach a plateau, stabilizing at approximately 9% after the completion of the fifth iteration.

## 5  Conclusion

The internet's influence on the business landscape cannot be overstated. It has revolutionized the way customers interact with businesses and seek out products and services worldwide. However, this vast digital realm also brings challenges, particularly in handling the overwhelming amount of information available. Online platforms like Twitter, Facebook, and Instagram have become powerful sources of customer feedback and reviews. These opinions are invaluable for businesses seeking to understand their customers better and improve their offerings. However, analyzing these reviews is no easy task, as they are often brief, noisy, and can lead to inaccurate text classifications.

In this research, this article presents a novel approach, the heterogeneous stacking ensemble-based text classification framework. The proposed framework involves a series of crucial steps, beginning with the preprocessing of textual documents to eliminate special characters, user handles, URLs, and stopwords. Tokenization is then applied to convert the preprocessed documents into individual word tokens, followed by integer encoding to represent them as sequences of integers with consistent lengths. To leverage the power of transformer models, the framework employs three different models: DistilBERT, ELECTRA, and BERT, generating three distinct sets of output predictions. The combination of these predictions into a single dataframe serves as the input to the stacking ensemble-based classifier model. Subsequently, the RoBERTa-based meta-level classifier is trained using these combined predictions. The experimental results, conducted on two diverse datasets, demonstrate the superior performance of the proposed stacking ensemble-based text classification analysis framework when compared to other existing frameworks in the literature. The achieved results highlight the framework's effectiveness in accurately classifying textual data.

Looking ahead, our future work aims to delve further into text classification analysis with transformer-based models, exploring improved utilization of word embeddings. By continuously refining and expanding the framework, we anticipate unlocking even greater potential in the field of text classification and sentiment analysis. In conclusion, the proposed heterogeneous stacking ensemble-based text classification framework presents a promising and robust solution for advancing text analysis tasks, with exciting prospects for practical applications across various domains.